\pdfoutput=1
\documentclass[11pt]{article}
\usepackage{authblk}
\usepackage[]{acl}
\usepackage{times}
\usepackage{latexsym}
\usepackage[T1]{fontenc}
\usepackage[utf8]{inputenc}
\usepackage{microtype}
\usepackage{amsmath}
\usepackage{amssymb}
\usepackage{array,etoolbox}
\usepackage{arydshln}
\usepackage{algorithm}
\usepackage[noend]{algpseudocode}
\usepackage{booktabs}
\usepackage{caption}
\usepackage{cleveref}
\usepackage{fancyhdr}
\usepackage{graphicx}
\usepackage{mathtools}
\usepackage{multirow}
\usepackage{microtype}
\usepackage{pifont}
\usepackage{pgfplots}
\pgfplotsset{compat=1.8}
\usepackage{relsize}
\usepackage{subcaption}
\usepackage{tikz}
\usepackage{tabularx}

\usepackage{url}
\usepackage{xcolor} 
\usepackage{bibentry}
\usepackage{xspace}
\def\impressions{\textsc{impressions}\xspace}

\def\findings{\textsc{findings}\xspace}

\title{Differentiable Multi-Agent Actor-Critic for Multi-Step Radiology Report Summarization}
\author[1]{\bf Sanjeev Kumar Karn}
\author[2]{\bf Ning Liu}
\author[3]{\bf Hinrich Sch\"{u}tze}
\author[1]{\bf Oladimeji Farri}
\affil[1]{Digital Technology and Innovation, Siemens Healthineers, Princeton}
\affil[2]{Corporate Technology, Siemens AG, Beijing}
\affil[3]{Center for Information and Language Processing (CIS), LMU Munich}
\affil[ ]{\tt \{sanjeev.kumar\_karn,oladimeji.farri\}@siemens-healthineers.com}
\affil[ ]{\tt liuning@siemens.com inquiries@cislmu.org}

\def\figlabel#1{\label{fig:#1}\label{p:#1}}
\def\figref#1{Figure~\ref{fig:#1}}
\def\eqref#1{Eq.~\ref{eqn:#1}}

\def\eqlabel#1{\label{eqn:#1}}

\def\tabref#1{Table~\ref{tab:#1}}
\def\tablabel#1{\label{tab:#1}\label{p:#1}}

\def\seclabel#1{\label{sec:#1}\label{p:#1}}
\def\secref#1{Section~\ref{sec:#1}}
\definecolor{forestgreen}{rgb}{0.13, 0.55, 0.13}

\newcommand\italicblue[1]{\textcolor{blue}{\textit{#1}}}
\newcommand\boldblue[1]{\textcolor{blue}{\textbf{#1}}}

\newcommand\Tstrut{\rule{0pt}{2.6ex}}  

\def\a{\textit{a}}
\def\m{\textit{m}}
\def\k{\textit{c}}

\def\algref#1{Algorithm.~\ref{alg:#1}}
\algdef{SE}[VARIABLES]{Variables}{EndVariables}
   {\algorithmicvariables}
   {\algorithmicend\ \algorithmicvariables}
\algnewcommand{\algorithmicvariables}{\textbf{global}}

\begin{document}
\maketitle
\begin{abstract}


The \impressions section of a radiology report about an
imaging study is a summary
of the radiologist's reasoning and conclusions,
and it also aids the referring
physician in confirming or excluding certain diagnoses. A
cascade of tasks are required to automatically generate an
abstractive summary of the typical information-rich
radiology report. These tasks include acquisition of salient
content from the report and generation of a concise, easily
consumable \impressions section. Prior research on radiology
report summarization has focused on single-step end-to-end
models -- which subsume the task of salient content
acquisition. To fully explore the cascade structure and
explainability of radiology report summarization, we
introduce two innovations. First,
we design
a two-step approach: extractive summarization followed by
abstractive summarization. Second, we additionally break down 
the extractive part into two independent tasks: extraction
of salient (1) sentences and (2) keywords. Experiments on 
English radiology reports from two clinical sites show our novel
approach leads to a more precise summary compared to
single-step and to two-step-with-single-extractive-process
baselines with an overall improvement in F1 score of 3-4\%.

\end{abstract}

\newcounter{notecounter}
\newcommand{\enotesoff}{\long\gdef\enote##1##2{}}
\newcommand{\enoteson}{\long\gdef\enote##1##2{{
\stepcounter{notecounter}
{\large\bf
\hspace{1cm}\arabic{notecounter} $<<<$ ##1: ##2
$>>>$\hspace{1cm}}}}}
\enoteson

\section{Introduction}
A diagnostic radiology report about an examination includes \findings in which the
radiologist describes normal and abnormal imaging results of
their analysis
\cite{dunnick2008radiology}. It also
includes \impressions or a summary that communicates
conclusions  about the findings and suggestions for the
referring physician; a sample report is shown
in \tabref{report_sample}. \findings are often lengthy and
information-rich. According to a survey of referring
physicians, \impressions may be the only part
of the report that is
read \cite{wallis2011radiology}. Overall, referring
physicians seem to appreciate the explainability (or
self-explanitoriness) of 
\impressions  as it helps them evaluate
differential diagnoses while avoiding additional
conversations with the radiologist or the need for repeat
procedures.

\begin{table}[!t]
\begin{center}
\begin{small}
\begin{tabular}{m{0.96\linewidth}}
\hline
\multicolumn{1}{c}{\findings}\\
\hline
\Tstrut \textsuperscript{\boldblue{$\psi$}}there is no evidence of \italicblue{midline shift} or \italicblue{mass effect}.\\
\Tstrut there is soft tissue swelling or hematoma in the right frontal or supraorbital region.\\
\Tstrut underlying sinus walls and calvarium are intact.\\
\Tstrut there is no obvious \italicblue{laceration}.\\
\Tstrut \textsuperscript{\boldblue{$\psi$}}there is subtle thickening of the \italicblue{falx} at the high convexity with its mid to posterior portion.\\
\Tstrut there is no associated subarachnoid hemorrhage.\\
\Tstrut \textsuperscript{\boldblue{$\psi$}}this likely reflects normal prominence of the \italicblue{falx} in a patient of this age.\\
\Tstrut \textsuperscript{\boldblue{$\psi$}}remote consideration would be a very thin \italicblue{subdural collection}.\\
\hline
\multicolumn{1}{c}{\impressions}\\
\hline
\Tstrut1) no definite acute intracranial process.\\
2) mild prominence of the falx is likely normal for this patient.\\
3) remote possibility of very thin subdural collection has not been entirely excluded.\\
\hline
\end{tabular}
\end{small}
\end{center}
\caption{\tablabel{report_sample}
\findings (top) and \impressions (bottom) sections of a
radiologist's report. \boldblue{$\psi$} indicates a sentence
in \findings that overlaps with sentences
in \impressions. Italicized words in \findings are core concepts (e.g., disorder and procedure) that assist in answering clinical questions.} 
\end{table}

A well known end-to-end method for text summarization
is \emph{two-step}: extractive summarization followed by
abstractive summarization. For
instance, \citet{chen-bansal-2018-fast} initially train
extractive and abstractive systems separately and then use
the extractive system as an agent in a single-agent
reinforcement learning (RL) setup with the abstractive
system as part of the environment. Their extractive system
extracts salient sentences and the abstractive system
paraphrases these sentences to produce a summary. This summary
is in turn  used to compute the reward for RL
training. However, this single-agent setup often fails to
extract some salient sentences or it extracts irrelevant
ones, leading to the generation of
incomplete/incorrect \impressions. We hypothesize
that granular categories of core concepts
(e.g., abnormalities, procedures) can be leveraged for
generating more comprehensive summaries. Thus, a separate RL
agent is dedicated to the task of extracting salient
keywords (core concepts) in the two-step system. The
novelty in this approach is that the new, second agent can now
collaborate with the first one and the two can influence each
other in their extraction decisions.

Multiagent reinforcement learning (MARL) requires that an agent coordinate with the other agents to achieve the desired goal. MARL often has centralized training 
and decentralized execution 
\cite{foerster2016learning,kraemer2016multi}. 
There are several protocols for MARL training, such as
sharing parameters between agents and 
explicit \cite{foerster2016learning,foerster2018counterfactual,sukhbaatar2016learning,mordatch2018emergence}
or implicit \cite{tian2020learning} communication between
agents by using an actor-critic policy gradient with a
centralized critic for all
agents \cite{foerster2018counterfactual}. The aim of these
protocols is to correctly assign credits so that an agent
can deduce its contribution to the team's success. To train
our cooperative agents that extract salient sentences and
keywords, we propose a novel Differentiable Multi-agent
Actor-Critic (DiMAC) RL learning method. We learn
independent agents in an actor-critic setup and use a
communication channel to allow agents to coordinate by passing
real-valued messages. As gradients are pushed through the
communication channel, DiMAC  is
end-to-end trainable across agents.




The novelties in the paper are threefold:
\begin{itemize}
\item a summarization
system that leverages core concepts via keywords, refines
them and makes them the basis for more fine-grained explainability \item a
multi-agent RL (MARL) based extractive component for a
two-step summarization framework, \item a Differentiable
Multi-agent Actor-Critic (DiMAC) with independent actors
leveraging a communication channel for cooperation
\end{itemize}

The remaining paper is structured as follows. In Section 2,
we provide a detailed description of our two-step
framework. In Section 3, we introduce the DiMAC training
algorithm. In Section 4, we describe training data and
experiments. In Section 5, we discuss the results. In
Section 6, we discuss related work. In Section 7, we
present our conclusions.

\section{Model}
\textbf{Problem statement.} We design a two-step
summarization framework that takes 
the \findings ($F$) section of
a radiology report (consisting of a sequence of sentences)
and a set of keywords ($K$) as input and 
produces an \impressions ($I$) section (consisting of a
sequence of sentences).

In the first step of the framework, the two extractors
independently select words and sentences from \findings $F$ but also coordinate such that the selection of salient
words is followed by the selection of the sentence
comprising these words. In the next step, a seq2seq
abstractor paraphrases the selected sentences to generate \impressions $I$. \figref{dimac_overview} illustrates the
proposed framework. We refer to \tabref{algo_notations} for
basic notations used in this paper. We often combine
notations to indicate a framework component concisely.



\begin{figure*}[t!]
\centering
\includegraphics[width=1.\linewidth]{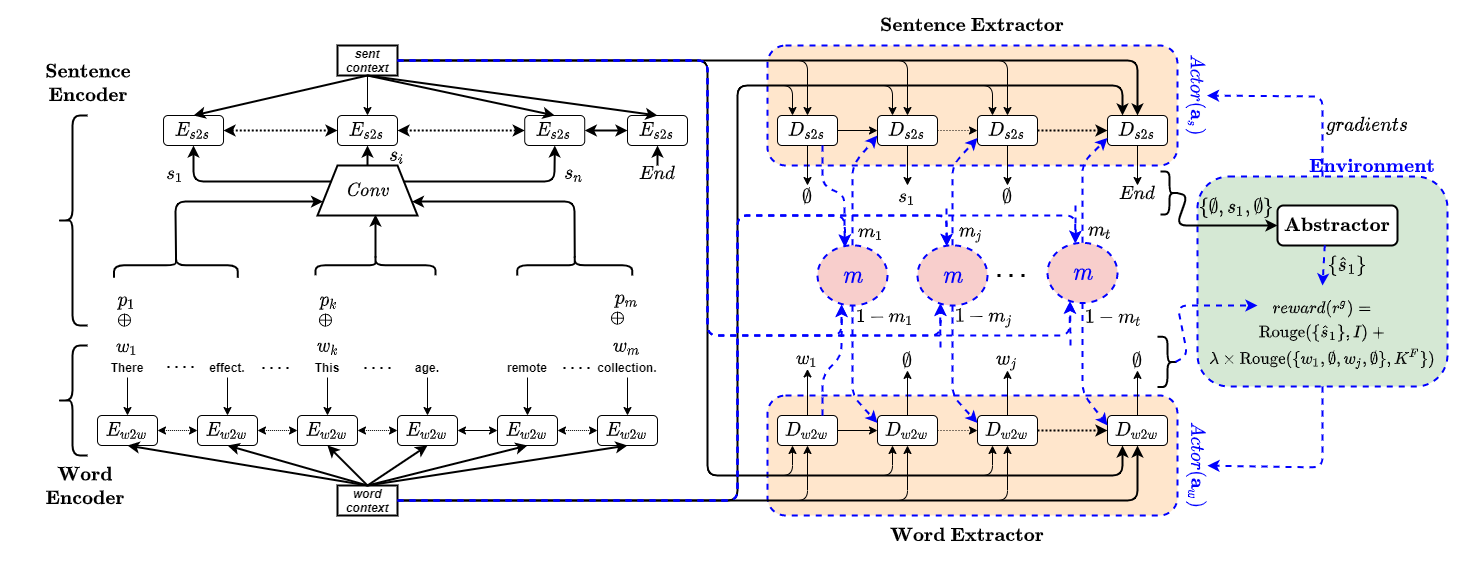}
\caption{Our two-step summarization framework.
DiMAC components (actors/extractors,
  communicator ($\m$), environment and communication between
  them) are indicated by blue dashed lines and arrows.
(i)
  The first
  step of the framework consists of encoder-extractor
  networks. Left side:
  sentence ($E_{s2s}$) 
and
  word
  ($E_{w2w}$) encoders.
  Right side: 
  sentence ($D_{s2s}$)
  and word ($D_{w2w}$) extractors.
 Word and sentence encoders are
  bi-directional LSTMs with word ($v^w$) and sentence
($h^s$)  embeddings  as input. A convolutional
  network ($\textit{Conv}$) obtains a sentence embedding ($h^s$) from word ($v^w$)
  and position ($v^p$) embeddings. An extractor is an LSTM
  pointer network with context vectors as input and either
  empty ($\emptyset$) or a source position as output at each
  step. (ii) In the second step of the framework, the seq2seq abstractor  paraphrases
  selected sentences. During DiMAC reinforcement learning, the communicator takes contexts and
  actor hidden states and sends them back messages
  ($m$). The critic is omitted. Abstracted sentences ($\hat{s}$)
  and selected words are used to compute
  rewards. {Figure best viewed in color}.
}\figlabel{dimac_overview}
\end{figure*}


\textbf{Two-step summarization framework.} The proposed
system includes encoder networks to encode words and
sentences into vector representations. It also includes two
pointer extractor networks \cite{NIPS2015_29921001} to
determine salient words and sentences by selecting their
indices. Both extractor networks run for the same number of
steps; however, at each step, the output index of one
extractor network is chosen while the other is set as empty
($\emptyset$). When the input is $\emptyset$, an extractor pauses its activity and guides the other extractor in an optimal direction.



\begin{table}[!t]
\begin{center}
\resizebox{1.0\linewidth}{!}{
\begin{small}
\begin{tabular}{p{0.03\linewidth}|p{0.25\linewidth}|p{0.08\linewidth}|p{0.29\linewidth}|p{0.03\linewidth}|p{0.33\linewidth}}
\hline
\multicolumn{6}{c}{\bf Notations}\\
\multicolumn{2}{c|}{\bf General}&\multicolumn{2}{c|}{\bf MLE}&\multicolumn{2}{c}{\bf RL}\\
\hline
$\textit{F}$&\findings&$\textit{E}$&\text{encoder network}&$\a$&\text{agent} (actors)\\ 
$\textit{I}$&\impressions&$\textit{D}$&\text{pointer network}&$\textit{c}$&\text{critic}\\
$\textit{K}$&{Keywords}&$\textit{w2w}$&\text{word LSTM}&$\textit{u}$&\text{action}\\
$\textit{w}$&\text{word}&$\textit{s2s}$&\text{sentence LSTM}&$\textit{m}$&\text{message value}\\
$\textit{s}$&\text{sentence}&$\alpha$&\text{attention score}&$\textit{r}$&\text{reward}\\
$\textit{p}$&\text{position}&$c$&\text{context vector}&$\textit{G}$&\text{discounted reward}\\
$\textit{m}$&\text{total words}&$\textit{v}$&\text{trainable vector}&$\textit{V}$&\text{value function}\\
$\textit{n}$&\text{total sentences}&$q$&\text{switch value}&$\textit{Q}$&\text{action value function}\\
$\textit{h}$&\text{hidden vector}&$\textit{W}$&{trainable matrix}&$\textit{A}$&\text{advantage function}\\
$y$&{train label}&$\textit{Conv}$&\text{CNN network}&&\\
\end{tabular}
\end{small}
}
\end{center}
\caption{\tablabel{algo_notations}
Notation used in this paper: general notation and notation
for two-step maximum likelihood estimation (\text{MLE}) and
reinforcement learning (RL). Notations are often combined,
e.g., $\textit{h}^{\textit{E}_{w2w}}$ refers to the word
encoder's hidden state vector and $\a_{w}$ to the word agent.}
\end{table}

\textbf{Encoder.} A bi-directional LSTM based word encoder,
$E_{w2w}$, is run on $m$ word embeddings of \findings
sentences to obtain word representations,
$\{h^{E_{w2w}}_{1},\cdots,h^{E_{w2w}}_{m}\}$. A
convolutional network ($Conv$) is run on concatenated word ($v^w$)
and position ($v^p$) embeddings in a sentence to obtain an
intermediate sentence representation ($h^s$). Then, a
bi-directional LSTM sentence encoder, $E_{s2s}$, leverages
the intermediate representations to obtain the final
sentence representations,
$\{h^{E_{s2s}}_{1},\cdots,h^{E_{s2s}}_{n}\}$.

\textbf{Extractors.}
Two LSTM based pointer extractors, i.e., word, $D_{w2w}$, and sentence, $D_{s2s}$, select a source word and sentence index at each step of decoding respectively. 
At any step $j$ of decoding, each extractor independently uses its hidden state $h^{D_{w2w}}_{j}$ and $h^{D_{s2s}}_{j}$ to compute an attention score over its source item $w_i$ and $s_k$ as:
\begin{equation*}
\begin{aligned}
&\alpha_{i,j}^{w}=\textrm{softmax}(\textit{v}^{T}\phi(\textit{W}^D h^{D_{w2w}}_{j}+\textit{W}^E h^{E_{w2w}}_{i}))\\&
\alpha_{k,j}^{s}=\textrm{softmax}(\hat{\textit{v}}^{T}\phi(\hat{\textit{W}}^D h^{D_{s2s}}_{j}+\hat{\textit{W}}^E h^{E_{s2s}}_{k}))
\end{aligned}
\end{equation*}
where $\textit{W}^D$, $\textit{W}^E$, $\textit{v}$, $\hat{\textit{W}}^D$, $\hat{\textit{W}}^E$ and $\hat{\textit{v}}$ are trainable parameters, $\textit{T}$ and $\phi$ are transpose and $\tanh$ functions respectively, and $\textrm{softmax}$ normalizes the scores. 
Word and sentence context vectors are computed using attention scores and encoder representations as $c^{w}_{j} = \sum_{i=1}^{m}{\alpha}_{i,j}^{w}h^{E_{w2w}}_{i}$ and $c^{s}_{j} = \sum_{k=1}^{n}{\alpha}_{k,j}^{s}h^{E_{s2s}}_{k}$ respectively. 

Additionally, at step $j$, the decision on whether word 
or sentence 
extractor output is set to $\emptyset$ is based on a switch
probability
$q_j=\sigma(\textrm{switch}({h}^{D_{w2w}}_{j},{c}^{{w}}_{j},{h}^{D_{s2s}}_{j},{c}^{{s}}_{j}))$,
where $\textrm{switch}$ is a feed-forward network (omitted
in \figref{dimac_overview}). The switch value of 0 or 1
indicates whether to set the output of sentence or word extractor to $\emptyset$.

Based on its current
cell state $h^{D_{s2s}}_{j}$,
$D_{s2s}$ computes the next cell state, both the context vectors
$c^{w}_{j}$ and $c^{s}_{j}$ and the selected source item
encoder representation, ${h}^{E_{s2s}}_{\cdot}$. Sharing
context vectors between extractors is similar to the cross
attention mechanism as described by
\citet{jadhav-rajan-2018-extractive}. In case $D_{s2s}$ is
at pause (i.e., $\textit{q}_j$=0), the $E_{s2s}$ end
representation is taken as the selected item
representation. $D_{w2w}$ follows the same approach to
compute its next state.

As we lack gold standard $\findings$ keywords and
sentence-wise one-to-one match between $\impressions$ and
$\findings$ to train networks to perform selection, we
heuristically obtain such labels. See
\secref{experiment} for details. We perform a maximum-likelihood (ML)
end-to-end training of the encoder-extractor networks to
minimize the following loss;
$\sum^{t}_{j=1}-(1-y^q_j)({y^w_j}\log
{\alpha^{w}_j})-y^q_j({y^s_j}\log {\alpha^{s}_k})-y^q_j\log
q_j$, where $t$ is the step when $D_{s2s}$ selects a dummy
$\textit{END}$, which indicates end of the extraction, and
$y^q_j$, $y^s_j$ and $y^w_j$ are heuristically obtained
switch, word and sentence selection labels at step $j$
respectively.


\textbf{Abstractor.}  The abstractor condenses each selected
sentence to a concise summary. We employ a pointer generator
network \citep{see-etal-2017-get} for this purpose. It uses
a copy mechanism to solve the out-of-vocabulary (OOV) problem
and a coverage mechanism to solve the repetition problem.
See \citep{see-etal-2017-get}
for  details. We independently train
the abstractor using heuristically obtained one-to-one matches
between $\findings$ and $\impressions$ sentences.




\section{DiMAC}
As extractor and abstractor are separately trained in a
two-step framework, \citet{chen-bansal-2018-fast} proposed
using RL training with the extractor assuming the agent role
and the abstractor as part of the environment to address the
separation. Furthermore, as RL loss is computed out of final
summary and ground-truth \impressions, RL training addresses
the error due to heuristic labels in the pre-trained
networks. Unlike \citet{chen-bansal-2018-fast}, our setup
involves multiple extractors, so we use MARL for the
coordination. In other words, the word and sentence
extractors ${D_{w2w}}$ and ${D_{s2s}}$ operate as RL agents
${\a_{w}}$ and ${\a_{s}}$ (\figref{dimac_overview}, right sidie).


In \cite{foerster2018counterfactual},
an actor-critic MARL has a centralized critic
and parameter-sharing actors. In contrast, our
extractors have different characteristics, e.g., amount of
selection (salient words greater than sentences) and size of
source representations; therefore, we exclude parameter
sharing between actors. Additionally, to not have actors
influence each other's policies, we have a critic that
estimates the value function by not conditioning on the actions
of other agents, thereby ensuring actor independence.
Furthermore, we introduce a communicator ($\m$) that
coordinates actors through message passing. 
The dedicated channel $\m$ addresses the issue of the
environment appearing non-stationary due to independent
agents; see
\cite{foerster2016learning,sukhbaatar2016learning,mordatch2018emergence}. The channel allows gradients to flow between actors,
transforming the setup into an end-to-end Differentiable
Multi-agent Actor Critic (DiMAC).  The actors and the
communicator are initialized with the maximum-likelihood
(ML) trained extractors and switch network, respectively.


\textbf{Actions.} Actors ${\a_{w}}$ and ${\a_{s}}$ have
action spaces of source words $\{w_{1},\cdots, w_{m}\}$
and sentences $\{s_1,\cdots, s_n\}$, respectively. At any
decoding step $j$, actors choose actions (i.e., source
selection) $\textit{u}^{\a_w}_j$ and $\textit{u}^{\a_s}_j$
by using policy networks $\pi^{\a_w}$ and
$\pi^{\a_s}$ and hidden states $h^{\a_w}_{j}$ and
$h^{\a_s}_{j}$. Due to the communication between actors in
DiMAC training, we intuitively expect some correlation in
the actions.



\textbf{Reward.}  For any decoding step $j$, if the communicator
indicates sentence selection ($\m=0$), a sentence reward
$r^{\a_s}_j$ is computed using $R_{1}$ (ROUGE unigram recall)
between the abstract summary $\hat{s}^{\a_s}_j$ of selected
sentence $s^{\a_s}_j$ (out of action
$u^{\a_s}_j$) from the abstractor and a ground-truth
\impressions sentence. We sequentially match summary and
\impressions sentences such that ${\a_{s}}$ learns to select
relevant sentences sequentially. Similarly, word reward
$r^{\a_w}_j$ for selected word $w^{\a_w}_j$ out of action
$u^{\a_w}_j$ is 1 if the word is in the subset of keywords
in \findings, $K^F$, else it is 0. Again, we match selected
and \findings keywords sequentially. When an agent
selects extra items, the reward for those selections is 0,
and thus, the agent learns to select only relevant sentences
and keywords.


In addition, joint actions of actors eventually generate a
global reward in a multi-agent cooperative setting as: $r^g$
= $R_{1}(\{\hat{s}^{\a_s}_{1},\cdots,
\emptyset,\cdots,\hat{s}^{\a_s}_{t}\}, I) + \lambda
R_{1}(\{w^{\a_w}_{1}, \cdots, \emptyset,\cdots,
w^{\a_w}_{t}\}, K^F)$, where $t$ is the step when $\a_s$
selects \textit{END} and $\lambda$ is a hyperparameter to
adjust the global word reward contribution. As $K^F$
keywords are not
gold-standard, we set $\lambda=0.1$; this  means that generated summary
sentences drive most of the global learning.  $r^g$
is included as the reward at the last step $t$ for both 
actors.



Action value functions ${Q}^{\a_w}_{j}$ and ${Q}^{\a_s}_{j}$ for actions $u^{\a_w}_j$ and $u^{\a_s}_j$
are estimated as $\mathbb{E}_{u^{\a_w}_{j:t},{h^{\a_w}_{j:t}}}[G^{\a_w}_j \mid h^{\a_w}_j,u^{\a_w}_j]$ and 
$\mathbb{E}_{u^{\a_s}_{j:t},{h^{\a_s}_{j:t}}}[G^{\a_s}_j \mid h^{\a_s}_j,u^{\a_s}_j]$, respectively, 
where $G^{\a_w}_j$ and $G^{\a_s}_j$ are discounted rewards computed as $\sum_{l=0}^{t-j}\gamma^{l}r^{\a_w}_{j+l}$ and $\sum_{l=0}^{t-j}\gamma^{l}r^{\a_s}_{j+l}$ and $\gamma=0.99$ is a hyperparameter.



\textbf{Critic.} Like the actors, the critic $\k$ is an
LSTM based network. It runs for the same number of steps as
the actors and estimates gradients to train them.  As the critic
is used only in training, at each step $j$, the critic
conditions on the actors' ground-truth selection indices,
${y}^{s}_{j}$ and ${y}^{w}_{j}$, as the actions and
uses these indices to obtain word and sentence
encoder representations. In addition to source representations, it
uses its state, $h^{\k}_{j}$, and attends to all  encoder
states, $\{h^{{E}_{w2w}}_{1},\cdots\}$ and
$\{h^{{E}_{s2s}}_{1},\cdots\}$) to estimate a value function
${V}_{j}$. ${V}_{j}$ is then used to compute advantage
functions $A^{\a_w}_j$ and $A^{\a_s}_j$ for actors as
${Q}^{\a_w}_{j}-{V}_{j}$ and ${Q}^{\a_s}_{j}-{V}_{j}$.
At any step, one of the two ground-truth actions
${y}^{s}_j$ / ${y}^{w}_j$  is empty. Therefore, the
computed value and action-value functions $V_j$ and ${Q_j}$
at that step intuitively become agent-specific, resulting in
independent agent learning. Finally, agent specific
advantage functions are used to compute actor gradients as
$\nabla_{\theta^{\a_w}}\log{\pi}^{\a_w}_j A^{\a_w}_j$ and
$\nabla_{\theta^{\a_s}}\log{\pi}^{\a_s}_j A^{\a_s}_j$.
Importantly, value, action-value and advantage
can be calculated in a single forward pass of the actor and
critic for each agent. See appendix for details and proofs.


\textbf{Communication.}  The communicator $\m$
(\figref{dimac_overview}, red circles) passes messages
between the actors. Actor previous hidden states and contexts,
$\textit{h}^{\a_s}_{j}$, $\textit{h}^{\a_w}_{j}$, $\textit{c}^{s}_{j}$ and $\textit{c}^{w}_{j}$, are
fed to $\m$ and a sigmoidal $m_{j}$ is obtained. Value
$m_{j}$ is fed to ${\a_{s}}$ while $1-m_{j}$ is fed to
${\a_{w}}$.
The gradient of $m_{j}$ flows between
actors during backpropagation and provides rich training
signal that minimizes the learning effort.

See Algorithm 1 for
DiMAC training algorithm details.

\label{app:DiMACalgo}
\begin{algorithm}[!t]
\begin{small}
\caption{Differentiable Multi-Agent Actor Critic}\label{alg:dimac_algo}
\begin{algorithmic}[1]
\Procedure{Train-DiMAC}{}
\State Initialize parameters of actors ($\a_w$ and $\a_s$), critic ($\k$) \& communicator ($\m$) as $\theta^{\a_{s}}\coloneqq{D}_{s2s}$, $\theta^{\a_{w}}\coloneqq{D}_{w2w}$,  $\theta^{\k}\coloneqq{D}_{w2w}$ \& $\theta^{\m}\coloneqq{switch}$
\For {each training episode \textit{i}}
		\State step $j \gets 1$
    		\While {action ${u}^{\a_s}_{j} \neq \text{END}$}
    		\State compute actors \& critic states 
    		\State sample actions $u^{\a_s}_{j} \text{ \& } u^{\a_w}_{j}$ 
    		\State compute rewards $r^{\a_w}_{j}$ \& $r^{\a_s}_{j}$ for $u^{\a_s}_{j}$ \& $u^{\a_w}_{j}$
    		\State compute message $m_{j}$ \& value function ${V}_{j}$ 
    		\State $j \gets j+1$
    		\EndWhile
        \State compute global reward $r^g$
        \For {$\textit{j}$ = $\textit{t}$ to 1}
        \State compute discounted reward $G^{\a_s}_{j}$ and $G^{\a_w}_{j}$ 
        \State estimate action-value functions ${Q}^{\a_w}_{j}$ \& ${Q}^{\a_s}_{j}$ 
        \State compute advantages ${A}^{\a_s}_{j} \text{ \& } {A}^{\a_w}_{j}$ 
        \State accumulate critic gradient $\Delta\theta^{\k}$
        \State accumulate actor gradients $\Delta\theta^{\a_s} \& \Delta\theta^{\a_w}$ 
        \EndFor
        \State update critic $\theta^{\k}_{i+1} = \theta^{\k}_{i}-\alpha\Delta\theta^{\k}$
        \State $\begin{aligned} & \text{update actors as }\theta^{\a_{s}}_{i+1} = \theta^{\a_{s}}_{i}+\alpha\Delta\theta^{\a_s} \text{ \& } \\&\theta^{\a_w}_{i+1}= \theta^{\a_w}_{i}+\alpha\Delta\theta^{\a_w}\end{aligned}$
\EndFor
\State \Return $\theta^{\a_{s}}$, $\theta^{\a_{w}}$ \& $\theta^{\m}$
\EndProcedure
\end{algorithmic}
\end{small}
\end{algorithm}


\definecolor{mypink2}{RGB}{219, 48, 122}
\definecolor{myblue1}{RGB}{172, 229, 238}
\definecolor{myred1}{RGB}{255, 133, 133}
\definecolor{mygreen1}{RGB}{0,255,0}
\definecolor{mywhite1}{RGB}{255,255,255}

\section{Experiments}\seclabel{experiment-data}

\begin{table}[t]
\footnotesize
\centering
\begin{tabular}{|l@{\hspace{0.1cm}}l|r|r|}
\hline
 & &\findings & \impressions \\
\hline
\#w &per sentence  & 10.54 (06.53) & 8.52 (05.80) \\
 \#s &per report & 8.23 (04.68) & 1.75 (01.16) \\
\#w &per report  & 86.77 (64.72) & 14.89 (15.81) \\
\hline
\end{tabular}
\caption{\tablabel{lang-diff}
Dataset statistics: number of words/sentences per sentence/report.
Standard
deviation in parentheses.
}
\end{table}

\subsection{Dataset}
We preprocessed and filtered radiology reports from two medical centers in the USA (Courtesy of Princeton Radiology, Princeton and University of Colorado Health). 
\footnote{Sentences
split using Stanford
CoreNLP \citep{manning-EtAl:2014:P14-5}. The following reports
are excluded: (a) no \findings and/or \impressions;
(b) \findings has fewer than 3 words; (c) \findings has fewer
words or fewer sentences than \impressions. We replace
special tokens like numbers, dates and abbreviations
and used scispacy lemmatization.} The resulting dataset
comprises 37,408 radiology reports, which we randomly split
into training (31,808), validation (3,740) and test sets
(1,860). \tabref{lang-diff}
gives dataset statistics.

\subsection{Experimental Setup}\seclabel{experiment}

\textbf{Training labels.}
Given an \impressions sentence, we find a unique \findings
sentence with the highest sentence similarity score. We
follow \citet{chen-bansal-2018-fast} and \citet{liu-lapata-2019-text} and use ROUGE-L as the sentence similarity scorer. Furthermore, they use a greedy matching algorithm that takes similarity scores of all \impressions and \findings sentence combinations and yields a sequence of unique \findings indices $\{y^s_1, \cdots\}$ of size equivalent to the length of \impressions. There is a 1-to-1 correspondence between \findings sentences at indices and \impressions sentences. We refer to the papers for more details. These 1-to-1 correspondence are used for abstractor pretraining.


We use AutoPhrase \citep{shang2018automated} to extract
keywords from training reports automatically. We select only
high-quality keywords, $K$, and avoid too frequent ones as
these can bias the system to only perform keyword
selection. We implement an empirical threshold determined by
hyperparameter search experiments.\footnote{AutoPhrase ranks
keywords using a quality score based on frequency. The
threshold is set on this score.} We then find a subset of
keywords, $K^F$, in \findings $F$ and compile their indices
$\{y^w_1, \cdots\}$.

As  the two extractors run for the same number of steps, we
interleave the above sentence and word indices
$\{y^s,\cdots\}$ and $\{y^w, \cdots\}$ into one sequence.
In more detail,
given a sentence index, all keywords
indices within that sentence are placed in the sequence,
followed by its index. A binary switch variable $y^q$ (with
values 0 and 1)
distinguishes the index type in the sequence, i.e.,
index refers to sentence vs.\
keyword. Both extractors require, during a
decoding step $j$, training labels $y^s_j$ and
$y^w_j$; we set the value of ``non-available type'' as indicated
by $y^q_j$ to $\emptyset$. For example, when $y^q_j$ is 0,
$y^w_j$ is $\emptyset$. Overall, an element in the final
sequence is a tuple of $y^q$, $y^s$ and $y^w$ and provides
training labels for the switch, word and sentence extractor
networks. See Appendix A for details on the interleaving of indices.

\textbf{Hyperparameters.} Included in Appendix C.

\textbf{Evaluation  measure.} We follow standard practice
and evaluate the quality of generated \impressions  by comparing
against ground-truth \impressions using ROUGE
\citep{lin-2004-rouge}.

\begin{table*}
\footnotesize
\centering
\begin{tabular}{p{0.4\textwidth}|ccc}
\hline
\textbf{Models}& \textbf{ROUGE-1} & \textbf{ROUGE-2} & \textbf{ROUGE-L}\\
\hline
LexRank \citep{journals/corr/abs-1109-2128} & 27.33 & 14.78 & 29.8 \\
PTGEN \citep{see-etal-2017-get} & 39.82 & 17.35 & 38.04\\
PTGEN+Coverage \citep{see-etal-2017-get} & 41.22 & 19.61 & 40.87\\
\citet{zhang-etal-2018-learning-summarize} & 44.16 & 22.67 & 43.07\\
BERTSUMAbs \citep{liu-lapata-2019-text} & 49.82 & 41.02 & 49.39 \\
BERTSUMExtAbs \citep{liu-lapata-2019-text} & 52.70 & 43.21 & 52.19\\
BART \citep{lewis2019bart} & 41.23 & 29.02 & 40.02 \\
Sentence Rewrite \citep{chen-bansal-2018-fast} & 59.82 & 48.54 & 59.11\\
DiMAC& \bf62.65&\bf51.55&\bf61.06\\
\hline
\end{tabular}
\caption{\tablabel{eng-result}
Results for baseline methods and DiMAC on the test split of the medical reports. 
The experimental setup is the same for all methods, i.e.,
the same train/validation/test split of the medical reports was used. 
Additionally, as DiMAC is a multi-agent two-step system built on top of Sentence Rewrite \citep{chen-bansal-2018-fast} (a single-agent two-step setup), we keep abstractor and all hyperparameters except those specific to DiMAC the same for a fair comparison.  
All  ROUGE scores have a 95\% confidence interval
of at most $\pm$0.50 as calculated by the official ROUGE script.}
\end{table*}

\subsection{Baseline Models}

In this section we describe the baselines we compare our
 model against: a wide variety of extractive and
 abstractive systems.

\textit{Extractive systems}

\textbf{LexRank} \citep{journals/corr/abs-1109-2128} is a graph-based method for computing relative importance in extractive summarization.

\textit{Abstractive systems}

\textbf{PTGEN} \citep{see-etal-2017-get} introduces an encoder-decoder model that can copy words from the source text via pointing, while retaining the ability to produce novel words through the generator.

\textbf{PTGEN+Coverage} \citep{see-etal-2017-get} introduces
 a coverage mechanism to the original PTGEN model to avoid repetition.
 
 \textbf{\citet{zhang-etal-2018-learning-summarize}} provides an
 automatic generation system for radiology \impressions
 using neural seq2seq learning. The model
 encodes background information of the radiology study and uses
 this information to guide the decoding process.

Self supervised learning has recently gained popularity as
parameters of large models can be trained with little
to no labeled data. Pre-trained language models in which a
transformer encoder is trained to reconstruct the original
text from masked text, e.g., BERT \cite{devlin2018bert}, have
become an important component in recent summarization
models \citep{liu-lapata-2019-text,zhang2020pegasus,zaheer2020big}. \textbf{We
also present results from experiments using these
summarization models}. Additionally, we experimented with a
\textbf{pre-trained seq2seq model} which is learned using different
self supervised techniques to reconstruct the original text,
e.g., BART \cite{lewis2019bart}.


\textbf{BertSumExtAbs}
\citep{liu-lapata-2019-text}
is an encoder-decoder summarization framework that
adopts BERT as its encoder. BERT is replaced by
ClinicalBERT \citep{alsentzer-etal-2019-publicly} in all our
experiments as it is adapted for the medical domain. At the
first stage, a model with the BERT encoder accomplishes an
extraction task. Then, the trained BERT encoder and a
6-layered transformer \citep{vaswani2017attention} are
combined to form an abstractive system. As the encoder in
the abstractive system is pre-trained multiple times in
comparison to the decoder, two separate Adam optimizers
(each with different warm-up steps and learning rates) are
used during training. As the training is performed in two
stages, \textbf{BertSumExtAbs} serves as the two-stage
abstractive summarization system baseline for our
experiments.\footnote{We require hyperparameters somewhat
different from the standard setup due to
the small radiology report
data size. Hyperparameter tuning yielded the following values. Batch size and initial learning rate of
BERTSumExt are set to 16 and 5e-4, batch size
in BERTSumExtAbs is 8 and initial learning rates of BERT and
transformer decoder in BERTSumExtAbs are 0.0005 and 0.005.}
We also include results from
\textbf{BERTSUMAbs},
a
single-stage version in which
encoder and decoder are trained only on the abstractive
task.

\textbf{BART} \cite{lewis2019bart}
is a state of the art transformer-based seq2seq model
similar to BERTSUMAbs. However, unlike BERTSUMAbs's
fine-tuning of the encoder and denovo training of the
decoder, for BART, both encoder and decoder are only
fine-tuned.

\textbf{Sentence Rewrite} \citep{chen-bansal-2018-fast} is a
two-step summarization model that initially extracts and
then rewrites the sentences. This model serves as a two-step
single agent baseline system for our experiments.


\section{Results}

 In this section, we compare results from our model and
 various baselines using both automatic  and human
 evaluation.

\textbf{Automatic Evaluation.}
\tabref{eng-result} shows report summarization results of
 various models trained and tested on the same data. Our
 DiMAC model surpasses extractive-only and abstractive-only
 baselines, including LexRank and PTGEN+Coverage. It also
 outperforms the two-step single agent baseline model
 (Sentence Rewrite \citep{chen-bansal-2018-fast}) and the
 two-stage
 BERTSUMExtAbs \citep{liu-lapata-2019-text}. Besides the
 pre-trained encoder of BertSumExtAbs, which is an advantage
 compared to other baselines, a denovo training of a large
 size decoder with a relatively small number of radiology
 reports may have led to overfitting. This might explain the
 scores compared to the two-step systems. Furthermore, a
 highly sophisticated semi-supervised training of the
 encoder and decoder of BART-base resulted in lower
 performance compared to our model, despite the relatively
 larger size (100x) of BART. We hypothesize that
 pre-training mostly on a different domain text (e.g.,
 Wikipedia, Books Corpus and News) and fine-tuning on small
 data could have adversely affected BART's performance in
 our setting. The domain difference may also contribute to the relatively lower performance of BART-base versus BERTSUMExtAbs, thereby signifying the importance of pre-training with relevant domain text.

Moreover, DiMAC offers approximately 18 to
28\% performance gains
over \cite{zhang-etal-2018-learning-summarize}, a
single-step single-agent summarization system designed
specifically for the radiology domain. In our opinion, the
performance improvements observed with DiMAC are likely
driven by the extract-then-abstract mechanism combined with
auxiliary (and salient) information from keywords, which
mimics the actual reasoning process of radiologists.

It is important to note that our model supports user-level validation by linking the predicted \impressions sentences to sentences in \findings, making the results explainable to radiologists and referring physicians.

\begin{table}[t!]
\footnotesize
\centering
\begin{tabular}{|l|r|r|r|r|}
\hline
&  &  &  & \textbf{Gwet}\\
& \textbf{Win} & \textbf{Tie} & \textbf{Lose} & \textbf{AC1}\\
 \hline
 \multicolumn{5}{|c|}{DiMAC vs. Base model} \\
\hline
Overall quality &25.00&59.37&15.63&.305  \\
Factual correctness  & 12.50 &84.37 & 03.13&.711\\  
 \hline
 \multicolumn{5}{|c|}{DiMAC vs. Ground Truth} \\
\hline
Overall quality &25.00 &46.87 & 28.13&.082  \\
Factual correctness  & 21.87&53.13&25.00&-.080\\  
\hline
\end{tabular}
\caption{\tablabel{human-evaluation-eng}
Percentage of 16 radiology reports for which human evaluators
rated DiMAC better than (win), the same as (tie) or worse
than (lose) the base model and ground truth on overall
quality and factual correctness.
We also provide Gwet's Agreement Coefficient as a measure of 
agreement
between raters; values below 0.2
indicate poor agreement, values above 0.8 indicate
very good agreement.}
\end{table}

\textbf{Human Evaluation.}
To assess the overall quality and factual
correctness \citep{zhang2019optimizing} of the \impressions
generated by DiMAC, we obtained evaluations from two
board-certified radiologists. We randomly selected 16
radiology reports from the test set.  For each radiology
report, we presented to the evaluators  its \findings
and three (blinded) versions of the summary,
i.e., \impressions:
(1) the ground truth, (2) Sentence
Rewrite \citep{chen-bansal-2018-fast} and (3) DiMAC. As Sentence Rewrite has a similar two-step approach,
i.e., extract-then-abstract, we evaluate the
qualitative performance of DiMAC with Sentence Rewrite
as the base model (instead of BERTSUMExtAbs as it is a
two-stage single-step system and also had lower Rouge scores
compared to Sentence Rewrite).

We shuffled the three summaries such that the order cannot
be guessed. Each radiologist rated the summaries
on two measures 
in relation to the
\findings: (1) overall quality and (2) factual
correctness and completeness. For example, the phrase
``pleural effusions'' is a fact (or imaging finding); but
the phrase ``small bilateral pleural effusions'' is a more
precise description and should therefore have a better
overall quality score. For each measure, we asked the
radiologists to score the summary as 1, 2 or 3
for bad, borderline or good.
Then we combined the assigned scores under
two comparisons: (1) our model versus the base model and
(2) our model versus ground truth.

We have 32 evaluations in total:
2 radiologists  $\times$ 16 reports. We compared the scores
provided by the radiologists to determine if
they were the same
(tie), higher (win) or lower (lose)  for our model vs.\
ground truth and our model vs.\ base model. 
\tabref{human-evaluation-eng} shows that
DiMAC has clearly better factual correctness than the base
model: 12.5\% of cases are better, 3.13\% are worse;
gwet AC1
\cite{gwet2008computing} inter-rater 
agreement for this result is strong.
DiMAC
exceeds the base model in 25\% (vs.\ 15.6\% ``lose'') of evaluations for overall
quality with moderate inter-rater agreement.
DiMAC is only slightly worse than ground truth in
overall quality (win: 25\%, lose: 28.13\%) and factual
correctness (win: 21.87\%, lose: 25\%) -- although
inter-rater agreement is low in this case.

\subsection{Qualitative Results Analysis} 
\tabref{topic_report}
shows a radiology report from our dataset (\findings
and \impressions) and \impressions generated by DiMAC and
the base model. Due to the
hierarchical connections between words and sentences, there
is significant overlap between the extracted sentences and
words. This phenomenon eventually contributes to the RL
sentence extraction reward and helps to extract sentences
with more keywords. The keywords include disease or clinical
diagnoses (e.g., nodule, lymphadenopathy, effusion),
anatomical concepts (e.g., hepatic) and qualifiers (e.g.,
recent, multiple, bilateral).
The baseline
model \citep{chen-bansal-2018-fast} erroneously states
``right greater than left
pleural effusions'', i.e., it hallucinates. In the sentence ``There is no axillary or
hilar lymphadenopathy'', the sentence reward is low and
eventually it is not extracted despite having the keyword
``lymphadenopathy''.

\begin{table}[t!]
\footnotesize
\centering
\resizebox{.99\linewidth}{!}{
\begin{tabular}{{p{7.7cm}}}
\hline
\\
\hline
\textbf{\findings from the report from a medical site}\\
\colorbox{myblue1}{There are \underline{\textbf{multiple bilateral}} lung\underline{ \textbf{nodules}} , most consistent}\\
\colorbox{myblue1}{with metastatic disease .}\\
There are more nodules on the right than the left .\\ An
enlarged prevascular lymph node measures 0.6 x 0.4 cm . \\
There is no axillary or
hilar \textbf{\underline{lymphadenopathy}} .\\
\colorbox{myblue1}{No pleural or pericardial \textbf{\underline{effusion}} is seen .}\\      
There is calcification in the aortic valve and coronary arteries .\\
\colorbox{myblue1}{There are numerous large \textbf{\underline{hepatic}} masses which have been}\\
\colorbox{myblue1}{better described on \textbf{\underline{recent}} ct scan of the abdomen .} \\

There is degenerative disease in the thoracic spine with mild compression of the superior endplate of a lower thoracic vertebral body .\\   
No suspicious osseous lesion is seen . \\
\hline
\textbf{\impressions from the report from a medical site}\\
 Multiple bilateral lung nodules , consistent with metastatic disease .\\
 Mediastinal lymphadenopathy .\\
 Multiple liver masses .\\
\hline
\textbf{\impressions generated by DiMAC}\\
Multiple bilateral lung nodules , most consistent with metastatic disease .\\
No pleural effusions .\\
Numerous hepatic masses , better described on recent ct scan of the abdomen .\\
\hline
\textbf{\impressions generated by base model}\\
Multiple bilateral lung nodules, most consistent with metastatic disease . \\
right greater than left pleural effusions .\\
enlarged right paratracheal lymph node .\\
numerous hepatic masses .\\
\hline
\caption{\tablabel{topic_report}
\findings and \impressions of a radiology report 
from the report from a medical site 
and  \impressions generated by  base model and DiMAC.
Extracted sentences are highlighted in blue. Extracted words are shown in bold and underlined. 
The base
model \citep{chen-bansal-2018-fast} erroneously states
``right greater than left
pleural effusions'', i.e., it hallucinates.
}
\end{tabular}
}
\end{table}

\section{Related Works}
\textbf{Abstractive Summarization.} An abstractive summary is a text consisting of novel phrases describing the content of the original text. Abstractive summarization involves a cascade of topic fusion and text generation \citep{hovy1999automated}. Each task in this cascade typically requires expert-derived annotations, which is labor-intensive and time-consuming. Thus, many recent abstractive summarization approaches focus on supervised/semi-supervised single-step end-to-end trainable models that implicitly address the sub-tasks of content acquisition and paraphrasing. 

As part of two-stage but single step abstractive
summarization, a pretrained encoder first learns the
extraction task independently. Then the pretrained encoder
is embedded into an encoder-decoder abstractive summarization
model to assist in better referencing the source content,
e.g., \citet{liu-lapata-2019-text,P18-1013}. On the other
hand, in two-step abstractive summarization, extractive
summarization is followed by abstractive summarization and
is trained end-to-end, e.g.,
\citet{chen-bansal-2018-fast}. Contrary to the two-stage
single-step approach, both extractive and abstractive
summarization are pretrained (and function) separately in a
two-step approach; however, an RL-based end-to-end training
enables alignment between them to generate better summaries.
DiMAC is a two-step abstractive system. 


\textbf{Multi-agent Reinforcement Learning (MARL).} In a
single-agent actor-critic
\citep{sutton1999policy,konda2000actor} policy gradient
method, an agent policy $\theta^\pi$ is optimized by
following a gradient computed using a value function
estimated by a critic. 
The simplest MARL setup applies policy gradients
independently (each agent with its own actor and critic) and
thereby restricts each agent to learn only from its own
action history \citep{tan1993multi}. From each agent’s point
of view in this setting, the environment is not stationary
and therefore, the RL stationary environment assumption is
violated.

\textbf{MARL with communication or collaboration protocols.} 
\citet{foerster2018counterfactual} proposed counterfactual policy gradients, which is an actor-critic policy gradient that leverages a centralized counterfactual critic that estimates value function for each actor by using actions performed by the other agents.
However, unlike our setting, actors in
\cite{foerster2018counterfactual} are similar and share
parameters. Additionally, the parameter sharing scheme has
the limitation that the agents lack tighter
coordination. \citet{foerster2016learning},
\citet{sukhbaatar2016learning} and
\citet{mordatch2018emergence} proposed to tightly coordinate
independent agents
 rather than use a dedicated channel.
As incorporating an explicit communication channel mimics human (bidirectional) interactions, we design a similar Differentiable Multi-agent Actor-Critic (DiMAC) RL for our setup. In DiMAC, each agent selects one of its actions and communicates with the others at every point in time.  
Thus, the resulting joint action (influenced by the agents' communication) would aim to reach the desired (optimal) goal. In the future, we will experiment with more variations of MARL (such as counter-factual critic) and  transformer-based networks.


\section{Conclusion}
In this work, we introduce a novel extractive approach into
a two-step RL-based summarization task
(extractive-then-abstractive). This approach is a MARL
(rather than the traditional single-agent RL) which includes
a new agent that extracts salient keywords from the source
text and collaborates with an agent that extracts salient
sentences. We also present a Differentiable Multi-agent
Actor-Critic (DiMAC) learning method, a novel yet simple MARL
training for independent agents communicating via a
dedicated channel. We apply the proposed two-step
summarization model with DiMAC MARL training to English
radiology reports. Results from our experiments indicate,
 based on automatic
and human expert evaluations,
that the DiMAC summarization model can outperform existing
baseline models for text summarization. Our summarization model
generates the \impressions to reflect human-level inference
and actionable information (e.g., salient sentences and
keywords) towards supporting improved workflow efficiency
and better-informed clinical diagnosis based on medical
imaging findings.

\section*{Acknowledgments}
We thank Dr.\ Asik Ali Mohamed Ali and Dr. Abishek
Balachandran for qualifying radiology reports and anonymized
summaries for human evaluation. We also thank Jashwanth N B
and Siemens Healthineers supercomputing team for training
infrastructure. Furthermore, we thank the anonymous reviewers
for their valuable feedback.

\textbf{Disclaimer}. The concepts and information presented in this paper are based on research results that are not commercially available. Future commercial availability cannot be guaranteed.

\bibliography{anthology,dimac}
\bibliographystyle{acl_natbib}



\appendix
\section*{Appendix}
\section{Training Labels}
In any training episode, we use Rouge-L and compute similarity scores between sentences in \findings and \impressions. Then, for each \impressions sentence, we find the \findings sentence that has the highest similarity score, and we compile its index. Furthermore, index compilation is a selection without replacement process, i.e., each sentence will only be selected once. This yields a sequence of unique sentence indices $\{y^s_1,\cdots\}$ of a size equivalent to the length of \impressions. Additionally, we flatten \findings sentences $\{s_1, \cdots, s_n\}$ to a long sequence of words $\{w_1, \cdots, w_m\}$. We then find words that are in the given keywords set $K$ and compile their indices $\{y^w_1,\cdots\}$. For example, in \tabref{report_sample}, salient sentence and word indices are $\{$1, 7, 8$\}$ and $\{$6, 7, 9, 10, $\cdots$, 81, 82$\}$ respectively. 

Finally, we interleave sentences and word indices $\{y^s_1,\cdots\}$ and $\{y^w_1,\cdots\}$ into one sequence to train extractors. Basically, given a sentence index, all keywords indices within that sentence are placed in the sequence. In addition, we use a binary switch variable ${y}^q$ (with values 0 and 1) to distinguish the index type in the sequence, i.e., ${y}^q$=0 implies sentence index and ${y}^q$=1 implies word index. Thus, the length of the binary switch variables sequence is the same as the interleaved indices. As extractors run for the same number of steps, training requires the labels $y^s_j$ and $y^w_j$ at any step $j$. However, the interleave sequence at any step includes only one out of the two. So, we set the value of "non-available type" as indicated by ${y}^q_j$ to $\emptyset$. Overall, an element in the final sequence is a tuple of ${y}^q$, $y^s$ and $y^w$. For \tabref{report_sample}, the final sequence of training labels is $\{$(1, $\emptyset$, 6), (1, $\emptyset$, 7), (1, $\emptyset$, 9), (1, $\emptyset$, 10), (0, 1, $\emptyset$), $\cdots$, (1, $\emptyset$, 81), (1, $\emptyset$, 82), (0, 8, $\emptyset$)$\}$.


\section{Encoder-Extractor Training}\label{sec:pretraining_appendix}
\algref{pretrain_algo} shows the training of word encoder (${E}_{w2w}$), sentence convolutional network ($\textit{Conv}$), sentence encoder (${E}_{s2s}$), word extractor (${D}_{w2w}$), sentence extractor (${D}_{s2s}$) and switch network (\textit{switch}).
\label{app:algo}
\begin{algorithm}[!h]
\begin{small}
\caption{Encoder-Extractor Training}\label{alg:pretrain_algo}
\begin{algorithmic}[1]
\Procedure{Train-Joint-Extractors}{}
\State \textit{Random Initialize}: ${E}_{w2w}, \textit{Conv}, {E}_{s2s}, {D}_{w2w}, {D}_{s2s}$ \& $\textit{switch}$
\For {1 to $\mid\textit{Reports}\mid$}
        \State $\{h_{1},\cdots,h_{n}\} \gets \textit{Conv}(\{s_1,\cdots,s_n\})$
		\State $\{h^{{E}_{s2s}}_{1},\cdots\} \gets {E}_{s2s}(\{h_{1},\cdots\})$
		\State $\{h^{{E}_{w2w}}_{1},\cdots,\} \gets {E}_{w2w}(\{w_{1},\cdots\})$
		\State $\textit{Loss} \gets Array()$
		\State $h^{{D}_{w2w}}_{1}, h^{{D}_{s2s}}_{1}\gets h^{{E}_{w2w}}_{m}, h^{{E}_{s2s}}_{n}$
		\For {$\textit{j}$ = 1 to ${t}$}
    		\State $\boldsymbol{\alpha}^{w} \gets \textit{Attn}(h^{{D}_{w2w}}_{j}, \{h^{{E}_{w2w}}_{1},\dots \})$
    		\State ${c}^{{w}}_{j} \gets \sum_{i=1}^{m}\alpha^{w}_i\times h^{{E}_{w2w}}_{i}$ 
    		\State $\boldsymbol{\alpha}^{s} \gets \textit{Attn}(h^{{D}_{s2s}}_{j}, \{h^{{D}_{s2s}}_{1},\dots \})$
    		\State ${c}^{{s}}_{j} \gets \sum_{k=1}^{n}\alpha^{s}_k\times h^{{E}_{s2s}}_{k}$
    		\State $q_{j} \gets \textit{switch}({h}^{D_{w2w}}_{j},{c}^{{w}}_{j},{h}^{D_{s2s}}_{j},{c}^{{s}}_{j}))$
    		\State $h^{{D}_{w2w}}_{j+1} \gets {D}_{w2w}(h^{{D}_{w2w}}_{j}, {c}^{{w}}_{j}, {c}^{{s}}_{j})$
    		\State $h^{{D}_{s2s}}_{j+1} \gets {D}_{s2s}(h^{{D}_{s2s}}_{j-1}, {c}^{{w}}_{j}, {c}^{{s}}_{j})$
    		\State $\textit{Loss}.\textsc{Add}(-(1-y^q_j)({y^w_j}\log {\boldsymbol{\alpha}^{w}}))$
    		\State $\textit{Loss}.\textsc{Add}(-y^q_j({y^s_j}\log {\boldsymbol{\alpha}^{s}}))$
    		\State $\textit{Loss}.\textsc{Add}(-y^q_j\log q_j)$
        \EndFor
        \State \text{compute gradients,} $\{\Delta_{{E}_{w2w}}$\textit{Loss}, $\cdots\}$
        \State \text{update} ${E}_{w2w}, \textit{Conv}, {E}_{s2s}, {D}_{w2s}, {D}_{s2s}$ \&  $\textit{switch}$
\EndFor
\State \Return ${E}_{w2w}, \textit{Conv}, {E}_{s2s}, {D}_{w2s}, {D}_{s2s}$ \& $\textit{switch}$
\EndProcedure
\end{algorithmic}
\end{small}
\end{algorithm}

\section{Hyperparameter}
We set the maximum limit for words in a report to 800 tokens, and the maximum number of sentences is truncated to 60 per report. We use word2vec \cite{mikolov2013efficient} on the training set to generate word embeddings of 128 dimensions.  The vocabulary is 50,000 most common words in the training set. The dimension of each intermediate sentence representation is 300 after using 1-D convolution filters with 3 different windows sizes (i.e. 3, 4, and 5). The dimension of all the LSTMs in our framework is 256. The optimizer used is Adam with a learning rate of 0.001 in the pre-training phase and 0.0001 in the RL training phase. We apply gradient clipping to alleviate gradient explosion using a 2-norm of 1.5. We adopt the early stopping method on the validation set. In the RL setting, the discounted factor $\gamma$ is set as 0.95. At test time, we use beam size 5 for beam search.

\section{Single Agent Actor-Critic}\label{sec:SAAC_appendix}
In the case of a single agent actor-critic RL, for any training episode, actor $\a$ uses its policy network $\pi^\a$ and samples actions $\{u^{\a}_1,\cdots,u^{\a}_t\}$ for $t$ time steps with each action $u^{\a}_k$ receiving a reward $r^{\a}_j$. Furthermore, at step $j$, a discounted reward is computed as $G^{\a}_{j} = \sum_{l=0}^{t-j}\gamma^{l}r^{\a}_{j+l}$. 

A batch of training episodes is used to estimate the actor's action value at step $j$ as ${Q}^{\a}_{j}=\mathbb{E}_{u^{\a}_{j:t},{h^{\a}_{j:t}}}[G^{\a}_j \mid h^{\a}_j,u^{\a}_j]$. Similarly, the critic ($\k$) estimates a value function for step $j$ as ${V}_{j}=\mathbb{E}_{h^{\a}_{j}}[G^{\a}_j\mid h^{\a}_j]$. An advantage function is computed as $A^{\a}_j={Q}^{\a}_{j}-V_j$. Policy gradient theorem computes the gradient to update the actor parameter $\theta^{\a}$ as
\begin{equation}
    \Delta\theta^{\a} = \mathbb{E}_{\theta^{\a}} \Bigg[\sum_{j=1}^{t}\nabla_{\theta^{\a}}\log{\pi}^{\a}_j A^{\a}_j\Bigg]
\eqlabel{SAAC_policy_gradient1}
\end{equation}

The value function component $V_j$ in the policy gradient helps to reduce the variance without changing the expectation as  
\begin{equation*}
\begin{aligned}
    -&\mathbb{E}_{\theta^{\a}} \Bigg[\nabla_{\theta^{\a}}\log{\pi}^{\a}_j V_j\Bigg]\\&
    = -\sum_{h}{d}^{{\pi}^\a_j}(h)\sum_{u^{\a}}\nabla_{\theta^{\a}}{\pi}^\a_j V_j\\&
    = -\sum_{h}{d}^{{\pi}^\a_j}(h)V_j\nabla_{\theta^{\a}}\sum_{u^{\a}}{\pi}^\a_j \\&
    = 0 
\end{aligned}
\eqlabel{SAAC_policy_gradient_1}
\end{equation*} where ${d}^{{\pi}^\a_j}(h)$ is the discounted ergodic state distribution \cite{sutton1999policy}. $V_j$ is a function of state and not action, thus moved outside $\nabla$, and since $\sum_{u^{\a}}{\pi}^\a_j$=1, the gradient becomes 0. $\Delta\theta^{\a}$ is empirically estimated using $N$ episodes in a training batch as
\begin{equation}
    \Delta\theta^{\a} \approx \frac{1}{N}\sum_{i=1}^{N} \Bigg[\sum_{j=1}^{t}\nabla_{\theta^{\a}}\log{\pi}^\a_j A^\a_j\Bigg]
\eqlabel{SAAC_policy_gradient2}
\end{equation}

\section{Multi Agent Actor-Critic}\label{sec:MAAC_appendix}
In the case of multi-agent actor-critic RL with a set of actors, $\textbf{\a}$=\{$\cdots,\a_{k},\cdots$\}, for any training episode, an actor $\a_{k}$ uses its policy network $\pi^{\a_k}$ and samples actions $\{u^{\a_k}_1, \cdots,u^{\a_k}_t\}$ for $t$ time steps with each action $u^{\a_k}_j$ receiving a reward $r^{\a_k}_j$. Furthermore, at step $j$, a discounted reward for $\a_{k}$ is computed as $G^{\a_k}_{j} =\sum_{l=0}^{t-j}\gamma^{l}r^{\a_k}_{j+l}$. 

Like the single agent actor-critic, a batch of training episodes is used to estimate the action value of $\a_k$ at step $j$ as ${Q}^{\a_k}_{j}=\mathbb{E}_{u^{\a_k}_{j:t},{h^{\a_k}_{j:t}}}[G^{\a_k}_j \mid h^{\a_k}_j,u^{\a_k}_j]$. 

The contribution of value function from a centralized critic at any step $j$ in the overall gradient is computed as 
\begin{equation*}
\begin{aligned}
    &-\mathbb{E}_{\theta^{\textbf{\a}}} \Bigg[\nabla_{\theta^{\textbf{\a}}}\log{\pi}^{\textbf{\a}}_j V_j\Bigg]
\end{aligned}
\eqlabel{MAAC_policy_gradient_1}
\end{equation*}
where $\theta^{\textbf{\a}}$ and ${\boldsymbol \pi}^{\textbf{\a}}$ are the actors' $\textbf{\a}$ joint parameters and policies respectively. $V_j$ is the value function computed by the critic at step $j$. We drop the step notation $j$ subsequently as all notations are specific to step $j$. The agent-wise break of policies and the contribution of the value function in the overall gradient is   
\begin{equation*}
\begin{aligned}
\\&
    = -\sum_{h}{d}^{{\pi}^{\textbf{\a}}}(h)\sum_{\a_{k}}\sum_{u^{\a_{k}}}\nabla_{\theta^{\a^k}}{\pi}^{\a_k}V\\&
    = 0
\end{aligned}
\eqlabel{MAAC_policy_gradient_2}
\end{equation*}
where ${d}^{{\pi}^{\textbf{\a}}}$ is the discounted ergodic state distribution, ${u}^{\a_{k}}$ is agent $\a_k$ action and $V$ is the estimated value function by the critic. Although two actors are running at each step in our DiMAC training, only one of them is active while the other is on pause ($\emptyset$ selection). Therefore, the contribution of the term $\sum_{\textbf{\a}}\sum_{u^{\a_{k}}}\nabla_{\theta^{\a^k}}{\pi}^{\a_k} V$ is similar to a single-agent scenario, and therefore, the gradient is 0. Furthermore, the critic estimated value ensures that the active agent gets rewarded for its action leading to the overall success.    


\end{document}